\documentclass[conference]{IEEEtran}
\IEEEoverridecommandlockouts
\usepackage{cite}
\usepackage{amsmath,amssymb,amsfonts}
\usepackage{algorithmic}
\usepackage{graphicx}
\usepackage{listings}
\usepackage{textcomp}
\usepackage{caption}
\usepackage{url}
\usepackage{xcolor}
\def\BibTeX{{\rm B\kern-.05em{\sc i\kern-.025em b}\kern-.08em
    T\kern-.1667em\lower.7ex\hbox{E}\kern-.125emX}}

\colorlet{punct}{red!60!black}
\definecolor{background}{HTML}{EEEEEE}
\definecolor{delim}{RGB}{20,105,176}
\colorlet{numb}{magenta!60!black}

\lstdefinelanguage{json}{
    basicstyle=\normalfont\ttfamily,
    numbers=left,
    numberstyle=\scriptsize,
    stepnumber=1,
    numbersep=1pt,
    showstringspaces=false,
    breaklines=true,
    captionpos=b,
    frame=lines,
    backgroundcolor=\color{background},
    literate=
     *{0}{{{\color{numb}0}}}{1}
      {1}{{{\color{numb}1}}}{1}
      {2}{{{\color{numb}2}}}{1}
      {3}{{{\color{numb}3}}}{1}
      {4}{{{\color{numb}4}}}{1}
      {5}{{{\color{numb}5}}}{1}
      {6}{{{\color{numb}6}}}{1}
      {7}{{{\color{numb}7}}}{1}
      {8}{{{\color{numb}8}}}{1}
      {9}{{{\color{numb}9}}}{1}
      {:}{{{\color{punct}{:}}}}{1}
      {,}{{{\color{punct}{,}}}}{1}
      {\{}{{{\color{delim}{\{}}}}{1}
      {\}}{{{\color{delim}{\}}}}}{1}
      {[}{{{\color{delim}{[}}}}{1}
      {]}{{{\color{delim}{]}}}}{1},
}

\definecolor{codegray}{gray}{0.95}
\definecolor{commentgreen}{rgb}{0,0.6,0}
\definecolor{keywordblue}{rgb}{0.2,0.2,0.8}
\definecolor{stringred}{rgb}{0.6,0,0}

\lstdefinestyle{pythonstyle}{
    backgroundcolor=\color{codegray},
    commentstyle=\color{commentgreen}\ttfamily,
    keywordstyle=\color{keywordblue}\bfseries,
    stringstyle=\color{stringred},
    basicstyle=\ttfamily\small,
    breaklines=true,
    frame=single,
    numbers=left,
    numberstyle=\tiny,
    stepnumber=1,
    numbersep=1pt,
    numberstyle=\scriptsize,
    showstringspaces=false,
    captionpos=b,
    language=Python,
    frame=lines,               
    rulecolor=\color{black}, 
    morekeywords={as}, 
    emph={dc, access_config, read_data, access_algo_training, access_predict, read_model, read_data_predict, apply_predict},
    emphstyle=\color{magenta},
}

\begin{document}

\title{AI-based modular \\ warning machine for risk identification in proximity healthcare\\
\thanks{The authors acknowledge the financial support of the ``DHEAL-COM - Digital Health Solutions in Community Medicine'' (CUP: D33C22001980001) under the   Innovative   Health   Ecosystem   (PNC)—National   Recovery  and  Resilience  Plan  (NRRP)  program  funded  by the Italian Ministry of  Health.}
}

\author{
    \IEEEauthorblockN{
    Chiara Razzetta\IEEEauthorrefmark{1}, 
    Shahryar Noei\IEEEauthorrefmark{3}, 
    Federico Barbarossa\IEEEauthorrefmark{4}, 
    Edoardo Spairani\IEEEauthorrefmark{5}, 
    Monica Roascio\IEEEauthorrefmark{6},
    Elisa Barbi\IEEEauthorrefmark{7},\\
    Giulia Ciacci\IEEEauthorrefmark{7}, 
    Sara Sommariva\IEEEauthorrefmark{2}, 
    Sabrina Guastavino\IEEEauthorrefmark{2}, 
    Michele Piana\IEEEauthorrefmark{2}\IEEEauthorrefmark{1}\thanks{Corresponding author (email: michele.piana@unige.it)}, 
    Matteo Lenge\IEEEauthorrefmark{7}, 
    Gabriele Arnulfo\IEEEauthorrefmark{6},\\ 
    Giovanni Magenes\IEEEauthorrefmark{5},
    Elvira Maranesi\IEEEauthorrefmark{4}, 
    Giulio Amabili\IEEEauthorrefmark{4}, 
    Anna Maria Massone\IEEEauthorrefmark{2}, 
    Federico Benvenuto\IEEEauthorrefmark{2}, 
    Giuseppe Jurman\IEEEauthorrefmark{3}, \\
    Diego Sona\IEEEauthorrefmark{3}, 
    Cristina Campi\IEEEauthorrefmark{2}\IEEEauthorrefmark{1}
    }
    \\
    \IEEEauthorblockA{\IEEEauthorrefmark{1}IRCCS Ospedale Policlinico San Martino, Genova, Italy}
    \\
   \IEEEauthorblockA{\IEEEauthorrefmark{2}MIDA, Dipartimento di Matematica, Università degli Studi di Genova, Genova, Italy.}
    \\
    \IEEEauthorblockA{\IEEEauthorrefmark{3}Fondazione Bruno Kessler, Trento, Italy}
    \\
    \IEEEauthorblockA{\IEEEauthorrefmark{4}Scientific Direction, IRCCS INRCA, Ancona, Italy}
    \\
    \IEEEauthorblockA{\IEEEauthorrefmark{5}Department of Electrical, Computer and Biomedical Engineering, Università di Pavia, Pavia, Italy}
    \\
    \IEEEauthorblockA{\IEEEauthorrefmark{6}Department of Informatics, Bioengineering, Robotics and System engineering (DIBRIS), \\Università degli Studi di Genova, Genova, Italy.}
    \\
    \IEEEauthorblockA{\IEEEauthorrefmark{7}Neuroscience and Human Genetics Department, Meyer Children's Hospital IRCCS, Firenze, Italy}
    }
    
\maketitle

\begin{abstract}
"DHEAL-COM - Digital Health Solutions in Community Medicine" is a research and technology project funded by the Italian Department of Health for the development of digital solutions of interest in proximity healthcare. The activity within the DHEAL-COM framework allows scientists to gather a notable amount of multi-modal data whose interpretation can be performed by means of machine learning algorithms. The present study illustrates a general automated pipeline made of numerous unsupervised and supervised methods that can ingest such data, provide predictive results, and facilitate model interpretations via feature identification.
\end{abstract}


\section{Introduction}

Risk analysis and population stratification are fundamental tasks across various application domains, including clinical, economic, and social contexts \cite{aven2015risk} . Traditional statistical approaches for risk analysis often rely on linear models and predefined parametric or semi-parametric scoring systems \cite{meeker2021statistical}. While these approaches have been successfully used to infer population-level characteristics, they show inherent limitations in making predictions regarding individuals. In recent years, machine learning approaches have emerged as a powerful alternative, offering greater flexibility and predictive accuracy across diverse domains \cite{hegde2020applications}. Modern machine learning methods typically incorporate large-scale, high-dimensional multi-modal data (including tabular data, time-series, images, and unstructural data, such as healthcare records) to generate data-driven stratification strategies. 

In this work, we focus on the implementation of modular pipelines for risk analysis within the scope of the ``DHEAL-COM - Digital Health Solutions in Community Medicine'' project, funded by the Italian Department of Health. This research and technology program aims to develop digital solutions to improve proximity healthcare at the national level, and involve the efforts of seven units organized according to a Hub and Spoke setting, which is building up a framework made of pre-existing multi-modal data, technological solutions for their generation, software tools for their exploitation and interpretation, high performance computing units for the execution of such tools. While previous studies have explored data-driven approaches in digital health, few have focused on scalable, interpretable pipelines integrating heterogeneous data types in community healthcare contexts. A survey conducted in collaboration with all the seven DHEAL-COM Spokes revealed the presence of:
\begin{itemize}
    \item Laboratory data.
    \item Clinical scores.
    \item Measurements from devices.
    \item Imaging data.
    \item Healthcare data.
\end{itemize}
We selected samples from each one of these data types and gathered the most relevant scientific questions intended to be addressed with this information. Specifically, we aimed at including in the system specific metrics for classification, regression, and clustering tasks \cite{pedregosa2011scikit}; feature selection techniques able to recursively remove features during training, while also providing a ranking of each feature’s importance \cite{chandrashekar2014survey,camattari2024classifier}; an automated approach for handling missing data \cite{emmanuel2021survey}; a process for tuning the hyper-parameters in the algorithms based on cross-validation \cite{probst2019tunability,montesinos2022overfitting}; and Synthetic Minority Oversampling TEchniques (SMOTEs) to generate synthetic examples when a better balancing of the training set is required \cite{chawla2002smote}. Of course, the general goals of the pipeline are to allow the users to both obtain predictive indications of the follow-up of the conditions and of their handling, and realize a model interpretation by understanding the influence of individual features on these predictions.

Outline of the paper is the following: Section II describes the software architecture that includes the input/output module, the implementation workflow, the machine learning algorithms, and the evaluation criteria for their performances. Section III illustrates the metrics used to assess the algorithms' performances and the model interpretation process. Our conclusions are offered in Section IV.

\section{Software architecture}

We proposed a modular approach, where each pipeline is composed of independent macro-blocks that can be assembled according to the specific needs of a given use case. Although this design introduces greater complexity during the initial development phase, it ensures better scalability and maintainability, allowing the pipelines to remain agnostic to the characteristics of individual datasets.

The system is entirely developed in Python \cite{van2007python}, a language widely adopted in the scientific community due to its rich ecosystem of libraries for machine learning and data processing. To support collaborative development and ensure proper version control, a GitHub repository is used (\url{https://github.com/dhealcom/wp4_pipeline}). Additionally, all system components are documented using the standard Python docstring format, which allows automatic generation of documentation in HTML and LaTeX through tools such as Sphinx. This ensures easy access to documentation for both end users and developers.

\subsection{Input ingestion}
To provide robust configurability and customization, both the operational parameters (see Listing \ref{lst:algoTrain_config}) and the data characteristics (see Listing \ref{lst:input_config}) are managed through JSON configuration files. 
Additionally, the reuse of pre-trained algorithms can also be managed through a JSON configuration file, further simplifying system integration and customization (see Listing \ref{lst:algoPredict_config}).

This choice is motivated by the increasing adoption of web-oriented tools such as JSON Forms, which allow the automatic generation of user interfaces based on the JSON schema \cite{pezoa2016foundations}. 
This approach facilitates user interaction with the system, enabling even non-technical users to accurately complete the required configuration files and integrate them into real-world applications. 

So far, tabular data are accepted but we are extending the data input module in order to manage other kinds of medical data, like images and time series (see \ref{sec:conc}).

\begin{lstlisting}[language=json,firstnumber=1, basicstyle=\small\ttfamily, caption={Example of data reading configuration file.},
label={lst:input_config}]
{
 "services": {
    "log_prefix": "log"
    },
 "runtime": {
    "run_id": 781
    },
 "dataset": {
    "name": "test",
    "type": "point-in-time",
    "_comment": "others: longitudinal, time-series, imaging",
    "format": "xlsx",
    "_comment": "csv,  xlsx"
    },
 "group": "Type",
 "PatientID": "Sample",
 "labels": [ "Type"],
 "time": "",
 "features2drop": ["A", "B"],
 "phase": "training_predict",
 "_comment": "training, training_predict",
 "categorical_features": ["C","D"],
 "split_percentage": 80,
 "split_type": "random",
 "_comment": "random, sequential, iterative"
}
\end{lstlisting}

\begin{lstlisting}[language=json,firstnumber=1, basicstyle=\small\ttfamily, caption={Example of algorithm configuration for the training phase.},
label={lst:algoTrain_config}]
{
  "algorithm": {
    "phase": "training",
    "config_name": "AggClustering",
    "description": "AggClustering",
    "type": "clustering",
    "parameters": {
      "preprocessing": {
        "standardization_feature": true,
        "standardization_label": false,
        "scaling_feature": false,
        "scaling_label": false
      },
      "data_inputation":{
        "perc_nan_to_drop": 0.5,
        "categorical": "most_frequent", 
        "_comment": "random, most_frequent",
        "not_categorical": "mean", 
        "_comment": "mean, median, regression"
      },
      "AggClustering": {
       "n_clusters":5,
       "linkage": "average"
        }
      }
    }
}
\end{lstlisting}

\begin{lstlisting}[language=json,firstnumber=1, basicstyle=\small\ttfamily, caption={Example of an algorithm configuration for reusing a pre-trained model. The system uses the "name" field to determine the data reading parameters, and the "description" field to apply the appropriate algorithm settings.},
label={lst:algoPredict_config}]
{
 "services": {
	"log_prefix": "log"
	},
 "runtime": {
    "run_id": 781
    },
 "dataset": {
    "name":"test",
	"type": "point-in-time",
	"format": "xlsx",
	},
 "description": "AggClustering"
}
\end{lstlisting}

\subsection{Workflow}
The proposed package follows a modular approach with the goal of creating a tool that can be used as broadly as possible.

We can distinguish three phases:
\begin{itemize}
\item Configuration and data loading.
\item Configuration and training of the model.
\item Prediction on additional data.
\end{itemize}

\begin{figure}[htbp]
\centerline{\includegraphics[width = 0.5\textwidth]{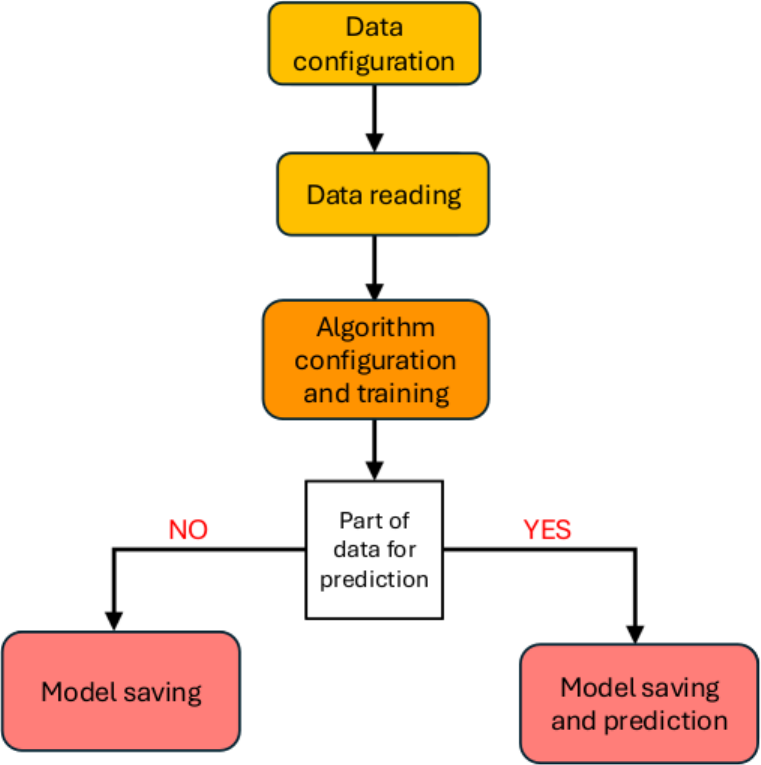}}
\caption{Complete pipeline of data loading, training, and prediction.}
\label{fig1}
\end{figure}

\begin{figure}[htbp]
\centerline{\includegraphics[width = 0.2
\textwidth]{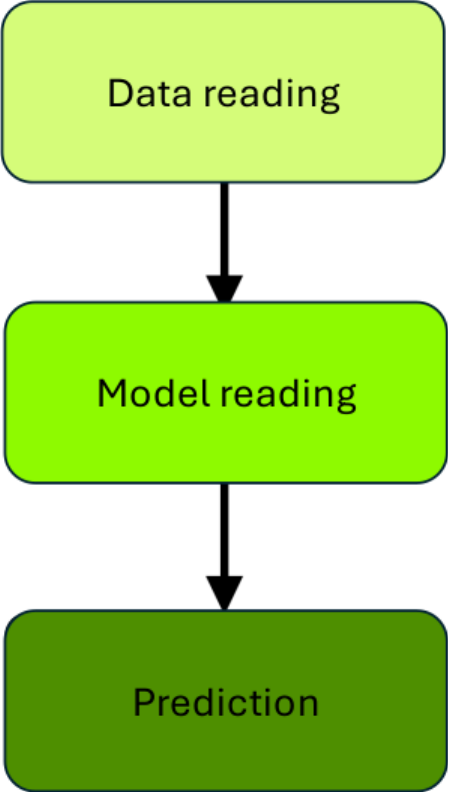}}
\caption{Pipeline for using a pre-trained model.}
\label{fig2}
\end{figure}

The possible workflows can be summarized with the two diagrams below: the first one (Figure \ref{fig1}) describes the complete pipeline of data loading, training, and prediction; the second one (Figure \ref{fig2}) outlines the few steps required to use a pre-trained model.
In general, only a few lines of code are needed to access the pipelines (see Listing \ref{lst:main}); the only requirement for training and reusing the models is to provide the correct configuration files.
\begin{lstlisting}[style=pythonstyle,firstnumber=1, basicstyle=\small\ttfamily, caption={Illustrative example of how to apply the proposed pipeline system for data loading, model training through JSON-based configuration, and prediction using either a freshly trained or a previously stored model.},
label={lst:main}]
    import DHEAL_COM_WP4_pipeline.DHEAL_COM_engine as dc

"""Configuration and data loading"""
input_folder = 'Data/'

input_file_data = 'example.xlsx'
cfg_input_data = 'input_configuration.json'

db = dc.access_config(cfg_input_data)
dd = dc.read_data(input_folder + input_file_data, db)


"""Configuration and training of the model"""

cfg_train = 'algorithm_configuration.json'
algo = dc.access_algo_training(cfg_train, dd, db)

"""Prediction on additional data"""

cfg_input_data_predict = 'predict_configuration.json'
input_file_predict_data = 'example_predict.xlsx'

if hasattr(dd,'df_predict'):
    predict = dc.access_predict(dd, algo)
else:
    rm = dc.read_model(cfg_input_data_predict)
    dd_predict = dc.read_data_predict(input_folder + input_file_predict_data, rm)
    predict = dc.apply_predict(rm, dd_predict)
\end{lstlisting}

\begin{figure*}[ht]
\centerline{\includegraphics[width = 1
\textwidth]{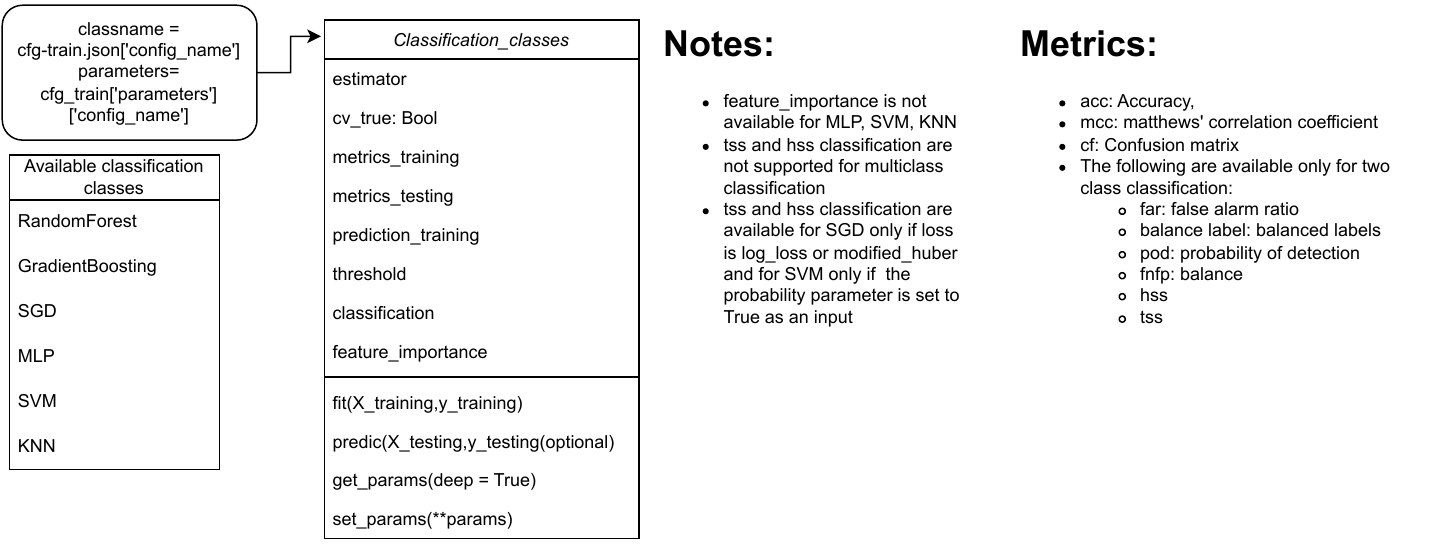}}
\caption{The Python class for the classification algorithms.}
\label{fig3}
\end{figure*}

%\begin{figure*}[ht]
%\centerline{\includegraphics[width = 1.3
%\textwidth, angle=90]{Classifier_methods.drawio.pdf}}
%\caption{The Python classifier method.}
%\label{fig4}
%\end{figure*}

\subsection{Implemented algorithms}

The pipeline utilizes algorithms for classification, regression, and clustering (see Table \ref{tab:algo}), and is also equipped with the following optional functions.
\begin{itemize}
\item A Recursive Feature Elimination (RFE) function. This feature selection technique allows features to be recursively eliminated during training, while simultaneously producing a ranking of the features used. If this technique is enabled in the algorithm's configuration file, all feature importance weights will be saved in the results. This function is now available for all algorithms that support feature weighting, enhancing the ability to identify the most relevant features for the model.
\item A module for handling missing data. Specifically, the system allows users to set a percentage threshold: if the percentage of missing data in a column is below the defined threshold, the column is removed; otherwise, the missing values are imputed according to the configured specifications. The configuration file allows selection of the most appropriate imputation technique based on the nature of the variable, with different modes for numerical and categorical variables. For numerical variables, the options include mean, median, or linear regression imputation; for categorical variables, it is possible to choose between random input or most frequent value.
\item A grid-search process based ion cross-validation for selecting the optimal (hyper)parameters.
\end{itemize}
Finally, to automatically address the problem of dataset imbalance, we introduced the option to enable Synthetic Minority Oversampling Technique (SMOTE).
SMOTE allows the generation of new synthetic examples for the minority class, increasing the amount of data in order to balance the class distribution.
It works by linearly interpolating existing examples from the minority class, preserving the original statistical properties of the dataset and avoiding the creation of simple duplicates.

\begin{table*}[ht]
    \centering
    \begin{tabular}{|c|c|c|c|}
    \hline
    Algorithm & Classification & Regression & Clustering \\ \hline\hline
Stochastic Gradient Descent (SDG) \cite{ketkar2017stochastic,patel2022global} & \checkmark & &\\ \hline
Elastic Net \cite{de2009elastic,tay2023elastic} & &  \checkmark &  \\ \hline
Gradient Boosting \cite{natekin2013gradient,bentejac2021comparative} & \checkmark & \checkmark &  \\ \hline
Random Forest (RF) \cite{RF,riello2025neuropsychological}  & \checkmark & \checkmark & \\ \hline
Multi-layerPerceptron (MLP) \cite{murtagh1991multilayer,piana2018flarecast}  & \checkmark & \checkmark & \\ \hline
Support Vector Machine (SVM) \cite{rosasco2004loss,noble2006support}  & \checkmark & \checkmark &  \\ \hline
K-Nearest Neighbors \cite{massone2000fuzzy,zhang2016introduction}  & \checkmark  &\checkmark & \\ \hline
K-Means \cite{likas2003global,taramasso2022distinct} & & & \checkmark \\ \hline
Agglomerative Clustering \cite{gowda1978agglomerative,tokuda2022revisiting}  & & & \checkmark\\ \hline
DBSCAN \cite{schubert2017dbscan}  & & &  \checkmark \\ \hline
    \end{tabular}
    \caption{List of available machine learning algorithms.}
    \label{tab:algo}
\end{table*}

\section{Evaluation criteria and interpretation of results}\label{sec:eval}
We implemented the more widely used evaluation metrics for each family of algorithms (see Table \ref{tab:metrics}). These metrics have been integrated in such a way that they are automatically calculated and saved during the training and prediction phases.

\begin{table*}[ht!]
    \centering
    \begin{tabular}{|c|c|c|c|}
    \hline
    Evaluation metric& Classification & Regression & Clustering \\ \hline\hline
Accuracy  & \checkmark & &\\ \hline
Precision  & \checkmark & &\\ \hline
Recall  & \checkmark & &\\ \hline
False alarm ratio& \checkmark & &\\ \hline
Probability of detection& \checkmark & &\\ \hline
Youden's index (True Skill Statistics - TSS)  & \checkmark & &\\ \hline
F1 score & \checkmark & &\\ \hline
Cohen's K (Heidke Skill Score - HSS) & \checkmark & &\\ \hline
Cross entropy & \checkmark & &\\ \hline
Matthews correlation coefficient & \checkmark & &\\ \hline
Mean squared error &  & \checkmark&\\ \hline
Root mean squared error &  & \checkmark&\\ \hline
Mean absolute error &  & \checkmark&\\ \hline
R$^2$ score &  & \checkmark&\\ \hline
Silhouette score &  & & \checkmark\\ \hline
Adjusted mutual information score &  & & \checkmark\\ \hline
Adjusted Rand score  &  & & \checkmark\\ \hline
V-measure &  & & \checkmark\\ \hline
    \end{tabular}
    \caption{List of available evaluation metrics.}
    \label{tab:metrics}
\end{table*}
%\begin{itemize}
%\item Classification algorithms: accuracy, precision, recall score, F1 score, confusion matrix, class prediction error, Cohen’s kappa score, cross-entropy, Matthews correlation coefficient, false alarm ratio, probability of detection, HSS score, TSS score, balance.
%\item Regression algorithm: mean squared error, root mean squared error, mean absolute error, R² score.
%\item Clustering algorithm: silhouette score, adjusted mutual information score, adjusted Rand score, V-measure.
%\end{itemize}

Regarding result interpretation, the option to compute and save SHAP values associated with the model has also been added.
SHapley Additive exPlanations (SHAP) is a tool based on Shapley value theory. SHAP generates values that represent the contribution of each feature to a given prediction, helping to identify potential biases or issues in the data \cite{sundararajan2020many}.
Specifically, SHAP provides not only a feature ranking, but also insights into which values (e.g., high/low for numerical variables, or specific categories for categorical variables) of each feature contribute most to the prediction.

To ensure reusability and seamless integration into broader web-based ecosystems, the trained models produced within the project are distributed along with their corresponding configuration files. 
This strategy supports consistent deployment by providing all necessary metadata and operational parameters in a structured and interpretable format. 
The results produced by the execution of the processing pipelines are systematically stored in JSON format, in line with the project's broader adoption of web-oriented standards (see Listing \ref{lst:out_training} and \ref{lst:out_pred}). 
This approach not only facilitates the reuse of models and reproducibility of experiments, but also enables straightforward integration with automatically generated user interfaces, thus supporting user-friendly access and interaction even in complex application scenarios.
\begin{lstlisting}[language=json,firstnumber=1, basicstyle=\small\ttfamily, caption={Example of output saved for the training phase.},
label={lst:out_training}]
{
 "config_data": {
    "metrics_training": {
        "Type": {
            "ARI": 0.7678293712508332,
            "AMI": 0.6816964180012729,
            "v-score": 0.7282823769983646,
            "Silhouette": 0.2419337904573403
            }
        }
    },
 "description": "AggClustering"
}
\end{lstlisting}

\begin{lstlisting}[language=json,firstnumber=1, basicstyle=\small\ttfamily, caption={Example of output saved for the prediction phase.},
label={lst:out_pred}]
{
 "testing_set": {
    "Type": {
        "ARI": 0.7678293712508332,
        "AMI": 0.6816964180012729,
        "v-score": 0.7282823769983646,
        "Silhouette": 0.2419337904573403
        }
    }
}
\end{lstlisting}

\section{Future developments and conclusions}\label{sec:conc}
We are currently working at both the maintenance and further algorithmic developments of the pipelines. In particular, we are working at:
\begin{itemize}
\item Data imputation strategies for tabular data.
\item Time series ingestion and processing. In the healthcare domain, time series are the format for Electroencephalography (EEG) \cite{kaur2015review,sorrentino2017inverse}, Magnetoencegphalography (MEG) \cite{hamalainen1993magnetoencephalography,sorrentino2009dynamical}, Electrocardiogram (ECG) \cite{houssein2017ecg}, and Cardiotocography (CTG) \cite{salini2024cardiotocography}.
\item DICOM and NIFTI \url{https://nifti.nimh.nih.gov/} images ingestion and processing; these formats are the ones adopted for Magnetic Resonance Imaging (MRI), Computed Tomography (CT), and Positron Emission Tomography (PET) \cite{bertero2006inverse}. Among the processing that will be available, we are working on a segmentation pipeline \cite{paolucci2024three} and a radiomics feature extraction pipeline \cite{cama2023segmentation}.
\item Loading of existing, pre-trained models and application to user dataset.
\end{itemize}
The proposed approach aims to provide a solid foundation for the design of complex, reusable, and easily extensible analytical solutions, contributing to the development of robust and user-friendly computational tools for risk analysis. 

\bibliographystyle{IEEEtran}
\bibliography{IEEEexample}

\end{document}